\documentclass{article}
\usepackage{ijcai25}

\usepackage{times}
\usepackage{soul}
\usepackage{url}
\usepackage[hidelinks]{hyperref}
\usepackage[utf8]{inputenc}
\usepackage[small]{caption}
\usepackage{subcaption} 
\usepackage{graphicx}
\usepackage{amsmath}
\usepackage{amssymb} 
\usepackage{amsthm}
\usepackage{booktabs}
\usepackage{algorithm}
\usepackage{algorithmic}
\usepackage[switch]{lineno}
\usepackage{pgfplots} 
\pgfplotsset{compat=1.5} 

\DeclareMathOperator*{\argmax}{argmax}

\title{RL + Transformer = A General-Purpose Problem Solver}

\author{
Micah Rentschler$^1$
\and
Jesse Roberts$^1$\\
\affiliations
$^1$Tennessee Technological University\\
\emails
mrentschler@tntech.edu,
jroberts@tntech.edu
}

\begin{document}

\maketitle

\begin{abstract}
    What if artificial intelligence could not only solve problems for which it was trained but also \textit{learn} to \textit{teach itself} to solve new problems (i.e., meta-learn)? In this study, we demonstrate that a pre-trained transformer fine-tuned with reinforcement learning over multiple episodes develops the ability to solve problems that it has \textit{never encountered before}—an emergent ability called \textbf{In-Context Reinforcement Learning (ICRL)}. This powerful meta-learner not only excels in solving unseen in-distribution environments with remarkable sample efficiency, but also shows strong performance in out-of-distribution environments. In addition, we show that it exhibits robustness to the quality of its training data, seamlessly stitches together behaviors from its context, and adapts to non-stationary environments. These behaviors demonstrate that an RL-trained transformer can iteratively improve upon its own solutions, making it an excellent \textit{general-purpose problem solver}.
\end{abstract}

\section{Introduction}


Imagine a Mars mission in which a robot's appendage malfunctions and becomes unusable. An adaptive agent controlling the robot could relearn how to operate without the appendage, allowing the mission to be completed successfully. This kind of adaptability is observed in nature, where animals adjust to the loss of limbs~\cite{cully2014robots}. Developing artificial intelligence (AI) systems with similar adaptability and problem-solving abilities in real, non-stationary environments remains a significant challenge.


Reinforcement learning (RL) has had great successes solving stationary control problems~\cite{sutton2018reinforcement}. However, the application of RL to the real world has been fraught with challenges. Compared to humans, RL methods suffer from \textit{low sample efficiency}~\cite{tsividis2017human,duan2016rl2fastreinforcementlearning}, meaning that they require a vast number of interactions with the environment before learning an effective policy. This inefficiency arises because they begin \textit{tabula rasa}—without any prior knowledge of the environment—and explore a wide range of possible actions and states to gather enough information to improve performance.

In short, RL excels at providing great solutions to specific, invariant problems where ample experience can be gathered to aid the learning process. However, many real-world problems do not require a specialist; they need a generalist with the ability to adapt to an ever-changing environment after only a few experiences. 

\begin{figure}
    \centering
    \begin{subfigure}[b]{0.32\linewidth}
        \centering
        \includegraphics[width=\linewidth]{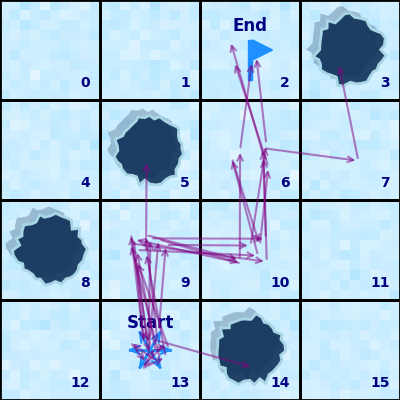}
        \caption{Early inference}
        \label{fig:subfig1}
    \end{subfigure}
    \hfill
    \begin{subfigure}[b]{0.32\linewidth}
        \centering
        \includegraphics[width=\linewidth]{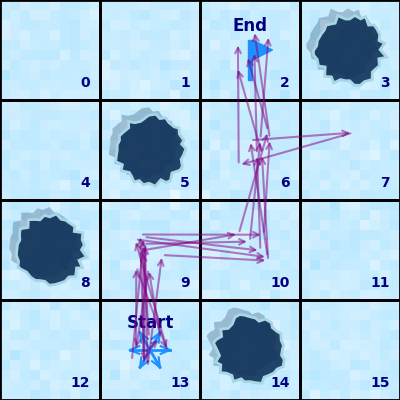}
        \caption{Mid inference}
        \label{fig:subfig2}
    \end{subfigure}
    \hfill
    \begin{subfigure}[b]{0.32\linewidth}
        \centering
        \includegraphics[width=\linewidth]{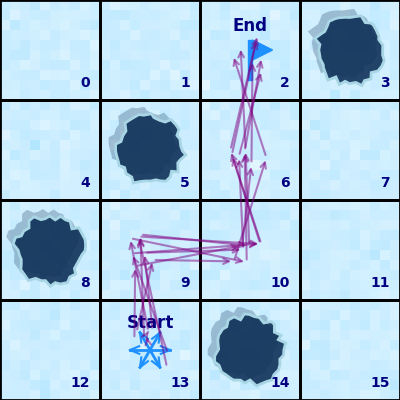}
        \caption{Late inference}
        \label{fig:subfig3}
    \end{subfigure}
    \caption{ICRL-trained Llama 3.1 learns to solve an unseen Frozen Lake environment. The trajectories in early (\subref{fig:subfig1}), mid (\subref{fig:subfig2}), and late (\subref{fig:subfig3}) interactions show solution refinement. Mistakes in early inference (e.g., falling into holes) disappear with experience in late inference.}
    \label{fig:frozen-lake-results}
    \vskip-1em
\end{figure}


Recent advances in other areas of AI have shown a remarkable ability to generalize. The introduction of the transformer architecture~\cite{vaswani2017attention} enables AI models to efficiently focus on relevant information, even when it is surrounded by a large amount of irrelevant context—a task that has been challenging for traditional fully connected neural networks. This capability has led to an emergent phenomenon called \textbf{in-context learning (ICL)}, as demonstrated by Brown et al.~\shortcite{brown2020language}, where the model learns to perform new tasks without adjusting its internal weights.

This ability to generalize and learn new tasks without retraining prompts us to ask: Is it possible to train a transformer to function as a reinforcement learning algorithm, improving its predictions based on a few experiences in its context without any additional weight updates? If so, can it generalize beyond its training data, learn with higher sample efficiency, and solve non-stationary environments—all \textit{without additional training}?


In this paper we show that: (1) Llama 3.1 8B can teach itself (meta-learn) through in-context experience by training on an RL objective, (2) the ability to meta-learn generalizes beyond the problem space in which it was acquired, (3) the model is robust to variations in the quality of its training data, (4) the model can assemble skills in a piecemeal manner to solve problems, and (5) the model learns to self-adapt when the environment changes.

This is the first work to show that a large language model (LLM) trained with the simple Deep Q-Network (DQN) algorithm~\cite{mnih2013playingatarideepreinforcement} can solve non-stationary problems such as adapting to the loss of a limb. Our results also raise thought-provoking questions: Should the Polyak averaging factor be increased in ICRL? Are parameter-efficient methods feasible and effective? And, what elements might be missing to enhance exploration? 

The successful integration of ICRL with transformer architectures signals a paradigm shift in RL research. By enabling transformers to meta-learn through in-context experiences, we pave the way for agents that can self-improve without additional training. This advancement represents a significant stride toward developing agents that are \textit{general-purpose problem solvers} able to navigate the complexities of the real world akin to living organisms.

\section{Background and Related Work}

Existing work (discussed below) has shown that RL-trained transformers are meta-learners. First, we review key works that combine transformers with RL, highlighting their contributions and limitations. Then we extend this literature, demonstrating that ICRL offers additional advantages. Specifically, ICRL-trained transformers possess the ability to combine learned skills in novel ways, learn effectively from suboptimal data, and adapt to non-stationary environments.

\subsection{Reinforcement Learning as Sequence Modeling}

A seminal effort in merging transformers with RL is the \textbf{Decision Transformer (DT)}~\cite{chen2021decision}, which reimagines RL as a sequence modeling problem. Instead of learning a policy or value function in the traditional sense, DT treats the return-conditioned trajectories as language tokens. The model is trained to predict the next action given a sequence of past states, actions, and the desired return (reward-to-go). By providing high target returns during inference, DT can generate action sequences that aim for high cumulative rewards. This approach demonstrates that transformers can model the dependencies in RL tasks without explicit policy optimization. However, DT relies heavily on the quality and diversity of its training data. Since it models the behavior implicit in the dataset, DT may struggle to generalize beyond the trajectories it has seen, lacking the ability to stitch together novel action sequences to achieve unseen goals. This dependence can limit its performance in environments requiring skill combination or in out-of-distribution scenarios.

To address these limitations, \textbf{Algorithm Distillation (AD)}~\cite{laskin2022context} proposes training transformers to imitate the behavior of RL algorithms themselves, rather than directly modeling action sequences. By learning from data generated by an RL algorithm, the transformer effectively distills the algorithm's policy into its weights. Remarkably, AD shows that transformers can be more sample-efficient than the algorithms they imitate, suggesting that they capture underlying patterns and structures facilitating efficient learning in a particular environment.

Building upon DT and AD, the \textbf{Decision-Pretrained Transformer (DPT)}~\cite{lee2024supervised} aims to overcome dependence on future rewards and specific algorithmic behaviors. DPT trains transformers to imitate an action oracle—a function providing optimal actions for given states—allowing the model to learn optimal policies directly. The authors demonstrate that DPT can perform RL effectively and sample-efficiently, possessing the ability to piece together partial trajectories to achieve goals, akin to dynamic programming. This trajectory stitching addresses a significant limitation of DT. However, DPT's reliance on an action oracle at train time necessitates prior knowledge of the optimal solution, limiting its applicability in settings where such an oracle is unavailable.

\subsection{Meta Reinforcement Learning}

Another avenue of research focuses on training transformers as in-context reinforcement learners, capitalizing on their sequence modeling capabilities and aptitude for meta-learning. Melo~\cite{melo2022transformers} demonstrated that transformers can be trained similarly to meta-RL algorithms like RL$^2$~\cite{duan2016rl2fastreinforcementlearning}. In this setup, the model adapts its policy based on the history of interactions within an episode. To address optimization instabilities often encountered in training transformers for RL tasks, they utilize the T-Fixup initialization~\cite{huang2020improving} to stabilize training. Their experiments reveal that transformers trained as in-context learners not only match but sometimes exceed the performance of traditional meta-RL methods. Notably, these transformers exhibit a degree of generalization to tasks slightly out of distribution, highlighting their capacity for rapid adaptation based on observed histories.

In parallel, researchers at DeepMind have introduced a transformer-based agent trained using meta-RL that adapts to solve complex tasks within timescales comparable to human learning~\cite{adaptiveagentteam2023humantimescaleadaptationopenendedtask}. Their agent demonstrates sample efficiency akin to humans, suggesting that transformers may employ learning strategies similar to those used by humans when confronting new challenges. This work underscores the potential of transformers as powerful meta-learners in RL settings.

\subsection{Unique Contributions}

Building upon the foundational works that merge transformers with reinforcement learning, we focus on harnessing the potential of ICRL. While previous studies have established that pre-trained transformers can be fine-tuned using reinforcement learning to solve novel problems~\cite{melo2022transformers,adaptiveagentteam2023humantimescaleadaptationopenendedtask}, our work advances this integration by uncovering and demonstrating several novel advantages of ICRL that have not been previously explored. Specifically, we show:

\begin{itemize}
    \item \textbf{In-Context Behavior Stitching}: ICRL-trained transformers can combine learned skills in novel ways to solve complex tasks. This ability indicates that the models have internalized principles akin to dynamic programming, allowing them to piece together previously acquired knowledge to tackle new challenges effectively.

    \item \textbf{Robustness to Low-Quality Data}: We find that ICRL reduces sensitivity to the quality of training data. Transformers trained using ICRL can learn effectively even from suboptimal actions, exhibiting strong generalization abilities despite imperfections in the training data.

    \item \textbf{Adaptation to Non-Stationary Environments}: Our experiments show that ICRL-trained transformers maintain high performance in changing environments by dynamically adjusting to new information. They prioritize recent interactions over outdated data, enabling them to adapt quickly to non-stationary settings and maintain robust performance.

\end{itemize}

These findings suggest that ICRL offers significant advantages in developing versatile AI systems capable of human-like adaptability.

\begin{figure*}
    \centering
    \includegraphics[width=\textwidth]{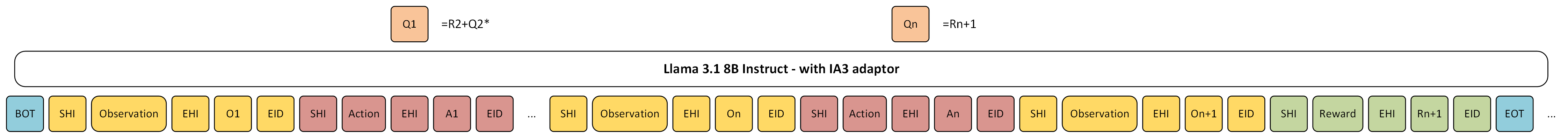}
    \caption{Fine-tuning LLaMA 3.1 8B Instruct with IA3 Adapters and a reinforcement learning objective. The model is fed sequences of states, actions, and (if nonzero) rewards, with every episode prefixed by the \texttt{<|begin\_of\_text|>} (BOT) token and terminated by the \texttt{<|end\_of\_text|>} (EOT) token. Tokens like \texttt{<|start\_header\_id|>} (SHI), \texttt{<|end\_header\_id|>} (EHI), and \texttt{<|eot\_id|>} (EID) separate the \textit{state}, \textit{action}, and \textit{reward}, mirroring how instruct models delineate \textit{user} and \textit{assistant} roles. The model predicts the Q-value of the current state for every action, updating the Q-values during training using the Bellman backup equation.}
    \label{fig:transformer}
\end{figure*}

\section{Methodology}

To explore the capabilities of \textbf{ICRL}, we employ the open-source \textbf{large language model (LLM)} called \textit{LLaMA 3.1 8B Instruct}~\cite{grattafiori2024llama3herdmodels}. We fine-tune this model using the \textbf{Deep Q-Network (DQN)} reinforcement learning algorithm~\cite{mnih2013playingatarideepreinforcement}, which enables the model to learn optimal actions through trial and error.

Our training data are collected from the parametric game \textit{Frozen Lake}~\cite{gymnasium2022}, a dynamic environment where the game parameters can be changed between episodes. Rather than focusing on solving a single, specific version of Frozen Lake, our objective is to enhance the model's performance across multiple episodes with varying game configurations. By doing so, we aim to improve the model's ability to generalize and find better solutions over time, thus highlighting the benefits of the ICRL approach.

This section aims to provide a clear understanding of our experiments and results. We begin by explaining the general problem formulation for a \textbf{Partially Observable Markov Decision Process (POMDP)}. Then, we review how reinforcement learning is applied to solve a POMDP. Next, we demonstrate how reinforcement learning can be applied to a pre-trained transformer model. Finally, we document our environment setup and data collection procedures.

\subsection{Partially Observable Markov Decision Process}

A \textbf{Markov Decision Process (MDP)} is a mathematical framework used to model decision-making problems where an agent interacts with a process whose next state depends solely on the previous state and action. In a \textbf{Partially Observable Markov Decision Process (POMDP)} the state is not fully observable. In such settings, the agent does not have direct access to the true state of the environment but must make decisions based on imperfect observations.

Formally, a POMDP is defined by the tuple \( (S, A, T, R, \Omega, O, \gamma) \), where:
\begin{itemize}
    \item \( S \) is a finite set of \textit{states} representing all possible configurations of the environment.
    \item \( A \) is a finite set of \textit{actions} available to the agent.
    \item \( T(s' \mid s, a) \) is the \textit{state transition probability}, the probability of transitioning to \( s' \) given action \( a \) in state \( s \).
    \item \( R(s, a) \) is the \textit{reward function}, the immediate reward received after taking action \( a \) in state \( s \).
    \item \( \Omega \) is a finite set of agent perceivable \textit{observations}.
    \item \( O(o \mid s', a) \) is the \textit{observation probability}, the probability of observing \( o \) after arriving at state \( s' \) and taking action \( a \).
    \item \( \gamma \in [0,1) \) is the \textit{discount factor} used to prioritize immediate rewards over future rewards.
\end{itemize}

Upon taking an action, the environment transitions to a new state \( s_{t+1} \) according to the transition probabilities \( T \). The agent receives a reward \( r_{t+1} \) given by the reward function \( R \) and observes the next observation \( o_{t+1} \) based on the observation probabilities \( O \).

If we define the trajectory up to time \( t \) as:
\begin{align}
    \tau_t = \{ o_0, r_0, a_0, o_1, r_1, a_1, \dots, a_{t-1}, o_t, r_t \}
\end{align}

In practice, the agent does not have access to the true state \( s \) at any time. Instead, it maintains a belief about the probability distribution over possible states given the history \( \tau_t \).

\subsection{Reinforcement Learning}

Reinforcement learning involves an agent interacting with an environment to maximize cumulative rewards over time. The agent observes the environment's state, takes actions, and receives rewards based on those actions.

We define the \textit{action-value function} (or \textit{Q-function}) \( Q^\pi(\tau, a) \) as the expected cumulative discounted reward obtained by taking action \( a \) given the history \( \tau \), and thereafter following a policy \( \pi \). In deep-RL the Q-function is normally a parameterized neural network denoted \( Q_{\theta} \) where \( \theta \) are the network's parameters. Formally, the action-value function is defined as:
\begin{align}
    Q^\pi_\theta(\tau, a) = \mathbb{E}_{a \sim \pi} \left[ \sum_{k=0}^{\infty} \gamma^{k} r_{t+k+1} \,\bigg|\, \tau_t = \tau, \, a_t = a \right]
\end{align}

The agent's objective is to find an optimal \textit{policy} \( \pi^* \) that specifies the best action to take based on the history, maximizing the expected cumulative discounted rewards. The optimal action-value function \( Q^*(\tau, a) \) corresponds to the maximum expected return achievable from history \( \tau \) by taking action \( a \) and thereafter following the optimal policy:
\begin{align}
    Q^*_\theta(\tau, a) = \max_{\pi} \; Q^\pi_\theta(\tau, a)
\end{align}

The optimal policy can be recovered for the optimal Q-function by taking the action that has the maximum value:
\begin{align}
\label{eq:policy}
    \pi^*_\theta(\tau_t) = \argmax_{a} \; Q^*_\theta(\tau_t, a)
\end{align}

An important property of \( Q^*_\theta(\tau, a) \) is that it satisfies a recursive relationship analogous to the \textit{Bellman optimality equation}~\cite{bellman1966dynamic}:
\begin{align}
    Q^*_\theta(\tau_t, a_t) = \mathbb{E} \left[ r_{t+1} + \gamma Q^*_\theta(\tau_{t+1}, a_{t+1}) \,\bigg|\, \tau_t, a_t \right]
\end{align}

Reinforcement learning algorithms aim to estimate \( Q^*_\theta(\tau, a) \) by iteratively applying this recursive relationship. A common approach is \textit{value iteration}, where the action-value target \( y \) is calculated:
\begin{align}
    y(\tau_t, a_t) = \mathbb{E} \left[ r_{t+1} + \gamma Q^*_\phi(\tau_{t+1}, \pi^*_\theta(\tau_{t+1})) \,\bigg|\, \tau_t, a_t \right]
\end{align}

The Q-network is trained by minimizing the loss:
\begin{align}
\label{eq:loss}
    L = \mathbb{E} \left[ \left( y(\tau_t, a_t) - Q_{\theta}(\tau_t, a_t) \right)^2 \right]
\end{align}

To facilitate fast convergence, several techniques are typically employed. Gradient flow through the target is stopped so that the current Q-value converges to the target while the target is fixed. However, because the target depends on the Q-network's own predictions from the previous iteration, this creates a moving target scenario. This can be mitigated by keeping a delayed copy of the Q-network (i.e. \( Q_\phi \)) from which we estimate the target and slowly update its parameters to follow the current Q-network (i.e. \( Q_\theta \))~\cite{vanhasselt2015deepreinforcementlearningdouble}. This process is called Polyak averaging
\begin{align}
\label{eq:polyak}
    \phi = \alpha*\theta + (1-\alpha)*\phi
\end{align}
and is controlled by a constant \( \alpha \).

Thus far, we have described the infinite horizon case. However, many games are episodic, so the action value is the expected sum of rewards until the game terminates, rather than extending to infinity. This is easily incorporated by defining the target function of the last action in a sequence to be equal to the reward alone.

By iteratively updating the Q-network parameters using optimization methods like stochastic gradient descent, the agent learns to approximate the optimal action-value function based on histories. This enables the agent to make informed decisions that maximize cumulative rewards, even in partially observable environments where the true state is not directly accessible.

In traditional reinforcement learning, only the last observation is provided to the network. However, when trying to induce the transformer to learn in-context, we provide the entire history of interactions. Thus, the transformer conditions its output on the whole trajectory \( \tau \). Only action tokens contribute to the training loss defined in Equation~\ref{eq:loss}.

\subsection{Transformer Network}

We selected \textit{LLaMA 3.1 8B Instruct} because it is a pre-trained transformer that has a demonstrated ability to perform ICL. We use an IA3~\cite{liu2022fewshotparameterefficientfinetuningbetter} adapter to decrease the computational load and memory requirements.

To train the network, we use a discount factor \( \gamma = 0.9 \) and scale the reward by multiplying it by \( 30 \). The delayed target adapter's weights are updated using Polyak averaging with a factor of \( \alpha = 0.1 \) (except when specified). Additionally, we use a learning rate of \( 1 \times 10^{-2} \), warmed from zero over the first ten batches, each consisting of 10 slices of data, with each slice being 4,096 tokens long.

During evaluation, as discussed in Section \ref{sec:exploration}, deploying the transformer without forced exploration results in poor performance. Thus, for each evaluation trial, we use an epsilon-greedy-style warm-up. In the first twenty episodes, epsilon (which represents the probability of using an action predicted by the transformer) is gradually increased from 0 to 1, which corresponds to gradually increasing the probability of letting the transformer choose the next action. When the transformer does not get to choose the action, we randomly select it from a uniform distribution. After twenty episodes, we let epsilon remain 1, so the transformer always chooses the next action.

\subsection{Environment}

The game \textit{Frozen Lake} is an excellent arena to demonstrate the power of ICRL. Frozen Lake has discrete states as well as discrete actions. We mapped the states to numbers corresponding to the tile numbers and mapped actions to the words \textit{up}, \textit{down}, \textit{left}, and \textit{right}. Our rationale was that LLaMA already understands these words and would be better able to adapt to the environment. Each parameterization of Frozen Lake can be represented by a map. The map designates certain starting point(s), end point(s), and hole(s) randomly placed throughout. The model is not allowed to see the map but must interact with the environment to learn it in order to maximize reward. When the player moves into a goal state, the player receives a reward of 1.0. For all other actions, the player receives a reward of 0.0.

\definecolor{color1}{HTML}{0B6E4F}
\definecolor{color2}{HTML}{2B4162}
\begin{figure}
    \centering
    \begin{tikzpicture}
    \begin{axis}[
        title={Unseen Environments},
        xlabel={Episode},
        ylabel={Cumulative Reward},
        xmin=0, xmax=30,
        ymin=0, ymax=1,
        xtick={0,10,20,30},
        ytick={0,0.2,0.4,0.6,0.8,1.0},
        legend pos=north west,
        ymajorgrids=true,
        grid style=dashed,
    ]
    \addplot[
        color=color1,
        line width=1.5pt,
        mark=square,
        mark size=1.5pt]
        table[col sep=comma,x=episode,y=mean_tau_0.1]{data/unseen.csv};
    \addplot[
        color=color2,
        line width=1.5pt,
        mark=triangle,
        mark size=2pt]
        table[col sep=comma,x=episode,y=mean_tau_0.01]{data/unseen.csv};
    \legend{\( \alpha = 0.1 \),\( \alpha = 0.01 \)}
    \addplot[
        <->,
        color=black,
        mark=none,
        line width=1.5pt,
        ] coordinates
           {(21,0.1) (21,0.9)};
    \node[anchor=west] at (axis cs: 22,.34) {900\%};
    \node[anchor=west] at (axis cs: 22,.27) {Increase};
    \end{axis}
    \end{tikzpicture}
    \caption{Mean cumulative reward over 50 trials as an ICRL-trained transformer improves its score on \textit{unseen} environments. Maps (i.e. environment parametrization) have never been trained on but are chosen from the same distribution as training examples. Significant improvement (approximately 900\% when \( \alpha = 0.1 \)) can be observed as the agent demonstrates that it has learned to solve unseen maps. Also, \( \alpha = 0.1 \) significantly  outperforms \( \alpha = 0.01 \).}
    \label{fig:in-results}
\end{figure}
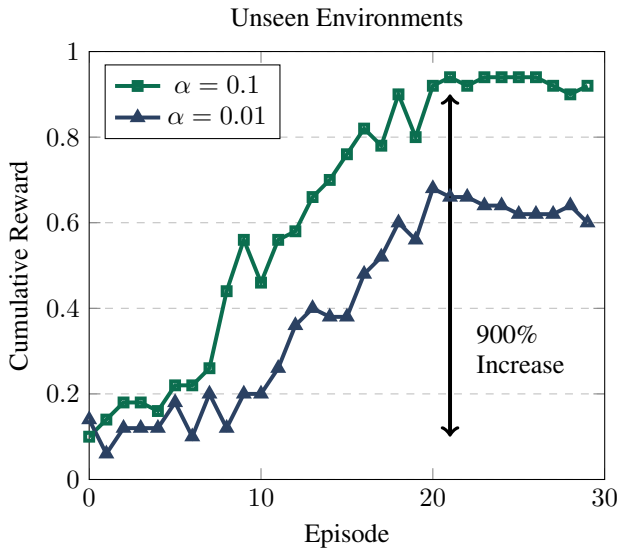

\subsection{Data}

Data are generated by training a traditional reinforcement learning algorithm on 250 different parameterizations of our chosen environment and collecting the data. It is extremely important to note that, unlike algorithmic distillation, we randomly mix episodes so that there is no inherent order to the data.

Our data are formatted in conversational form. Unlike the LLaMA instruct template, we use the roles of \textit{action}, \textit{observation}, and \textit{reward} instead of the roles \textit{user} and \textit{assistant} that are standard for instruct models.

We concatenate episodes of data together. Every 20 to 40 episodes, the environment parameterization (i.e., map in Frozen Lake) is changed so that the network can practice adapting to non-stationary environments. We call the 20 to 40 episodes with the same parameterization a \textit{set}. Multiple sets are combined together until 4,096 tokens are reached.

\section{Experiments}

In this section, we evaluate the performance of our ICRL-trained transformer across a variety of tasks to demonstrate its capabilities as a general-purpose problem solver. We examine its ability to solve both in-distribution and out-of-distribution examples, its capacity for in-context behavior stitching, its robustness to low-quality training data, and its adaptability to non-stationary environments. We also discuss the challenges associated with exploration in ICRL settings.

\definecolor{color1}{HTML}{0B6E4F}
\definecolor{color2}{HTML}{2B4162}
\begin{figure}
    \centering
    \begin{tikzpicture}
    \begin{axis}[
        title={Unseen Out-Of-Distribution Environments},
        xlabel={Episode},
        ylabel={Cumulative Reward},
        xmin=0, xmax=30,
        ymin=0, ymax=1,
        xtick={0,10,20,30},
        ytick={0,0.2,0.4,0.6,0.8,1.0},
        legend pos=north west,
        ymajorgrids=true,
        grid style=dashed,
    ]
    \addplot[
        color=color1,
        line width=1.5pt,
        mark=square,
        mark size=1.5pt]
        table[col sep=comma,x=episode,y=mean_tau_0.1]{data/out_of_distro.csv};
    \addplot[
        color=color2,
        line width=1.5pt,
        mark=triangle,
        mark size=2pt]
        table[col sep=comma,x=episode,y=mean_tau_0.01]{data/out_of_distro.csv};
    \legend{\( \alpha = 0.1 \),\( \alpha = 0.01 \)}
    \end{axis}
    \end{tikzpicture}
    \caption{Mean cumulative reward over 50 trials as an ICRL-trained transformer improves its score on unseen and out-of-distribution environments. Generated maps are larger than anything ever seen during training. Improvement can be observed (though not as significant as in the in-distribution case) as the agent demonstrates that it has learned useful behaviors even for environments outside the distribution of its training data.}
    \label{fig:out-results}
\end{figure}
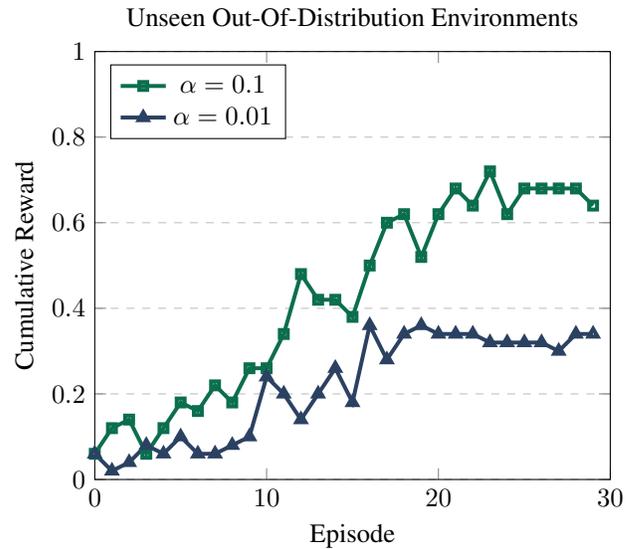

\begin{figure*}
    \centering
    \begin{subfigure}[b]{0.19\textwidth}
        \centering
        \includegraphics[width=\textwidth]{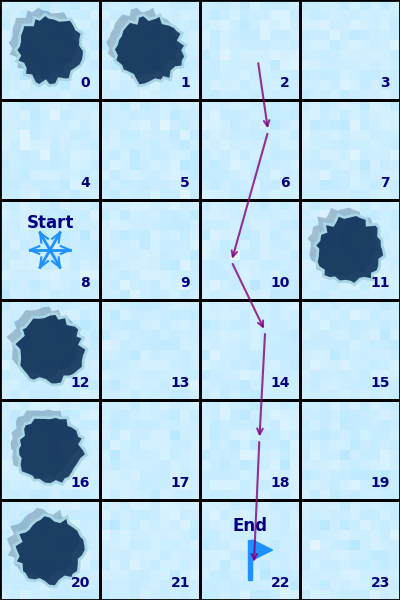}
        \caption{Example 1}
        \label{fig:chaining-subfig1}
    \end{subfigure}
    \hfill
    \begin{subfigure}[b]{0.19\textwidth}
        \centering
        \includegraphics[width=\textwidth]{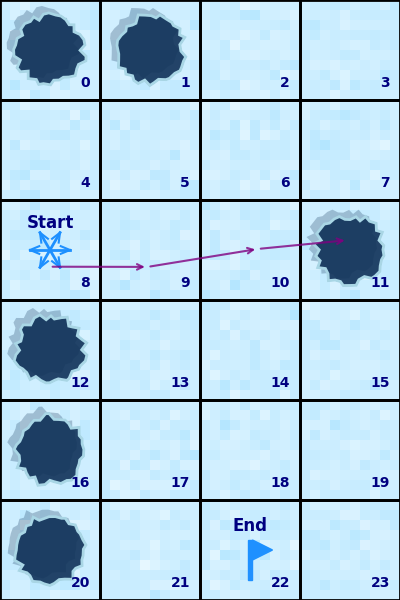}
        \caption{Example 2}
        \label{fig:chaining-subfig2}
    \end{subfigure}
    \hfill
    \begin{subfigure}[b]{0.19\textwidth}
        \centering
        \includegraphics[width=\textwidth]{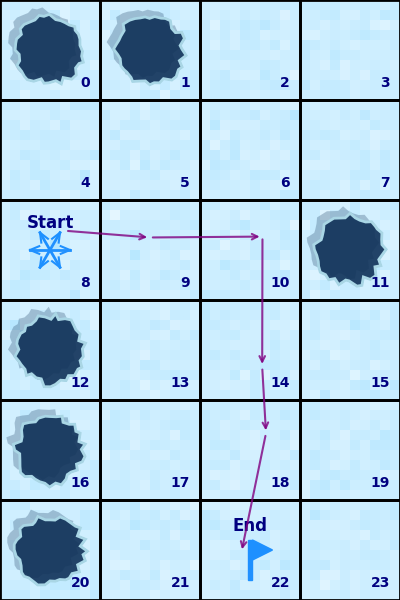}
        \caption{Trial 1}
        \label{fig:chaining-subfig3}
    \end{subfigure}
    \hfill
    \begin{subfigure}[b]{0.19\textwidth}
        \centering
        \includegraphics[width=\textwidth]{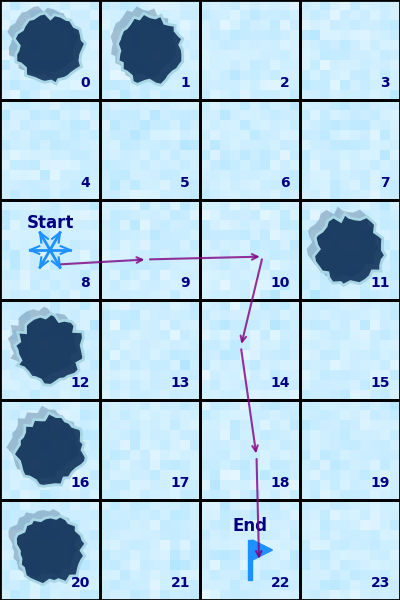}
        \caption{Trial 2}
        \label{fig:chaining-subfig4}
    \end{subfigure}
    \hfill
    \begin{subfigure}[b]{0.19\textwidth}
        \centering
        \includegraphics[width=\textwidth]{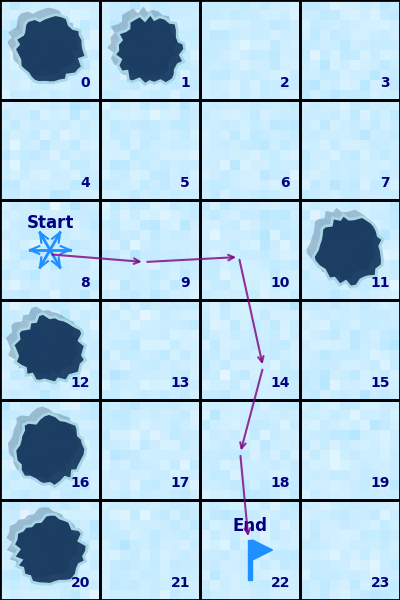}
        \caption{Trial 3}
        \label{fig:chaining-subfig5}
    \end{subfigure}
    \caption{Illustration of how ICRL combines different experiences to generate improved solutions. Subfigures (\subref{fig:chaining-subfig1}) and (\subref{fig:chaining-subfig2}) show two example trajectories provided as context to the ICRL transformer. The inference trials in subfigures (\subref{fig:chaining-subfig3}), (\subref{fig:chaining-subfig4}), and (\subref{fig:chaining-subfig5}) display the paths predicted by the transformer, which leverage information from both examples to develop an optimal solution.}
    \label{fig:chaining}
\end{figure*}

\subsection{Solving Unseen In-Distribution Examples}

To assess the transformer's ability to generalize to unseen but in-distribution examples, we evaluated its performance on new parameterizations of the Frozen Lake environment that were not included in the training set but were generated from the same distribution.

\textbf{Setup:} We generated 50 new Frozen Lake maps with widths and heights ranging from 3 to 5 tiles and randomly placed holes throughout the map, similar to the training data. The agent was not provided with any explicit map information and had to learn the optimal path solely through interaction with the environment. Each evaluation consisted of multiple episodes, allowing the transformer to learn and improve its policy through ICRL. We also examine the Polyak averaging constant's effect by testing both $\alpha=0.01$ and \( 0.01 \).

\textbf{Results:} As shown in Figure~\ref{fig:in-results}, the transformer effectively learned to navigate the new environments. During early episodes, the agent fell into holes, only reaching the goal 10\% of the time, but as it gained more experience, it achieved a 900\% improvement (when \( \alpha = 0.1 \)). This demonstrates that ICRL can successfully generalize to new, unseen maps within the same distribution, improving its performance over time without additional weight updates. We also notice a significant dependence on the Polyak constant, which suggests that allowing the target network to update quickly (i.e. \( \alpha = 0.1 \)) outweighs the benefits of increased stabilization (i.e. \( \alpha = 0.01 \)).

\textbf{Critical Observation:} As we explored specific failure examples, a pattern became clear... in the large majority of cases, the reason that the transformer fails to find the goal is that it fails to explore the whole map. We noticed that the agent almost always avoids holes but would get caught in loops where it traversed the same path repeatedly until it exceeded the number of allowable steps in Frozen Lake. We discuss this more in Section \ref{sec:exploration}.

\subsection{Solving Out-of-Distribution Examples}

To evaluate the model's ability to generalize beyond the training distribution, we tested it on Frozen Lake maps with configurations not encountered during training.

\textbf{Setup:} We created 50 out-of-distribution environments by using larger map sizes (e.g., widths and heights of 6 to 7 tiles). These maps are not only larger; they are also much harder, as a longer sequence of actions must be learned to reach the goal state. As before, the transformer had to learn to navigate these new and more complex environments solely through interaction. We also ablate the Polyak averaging constant by setting \( \alpha \) equal to \( 0.1 \) and \( 0.01 \).

\textbf{Results:} The transformer showed remarkable adaptability to these out-of-distribution environments. While performance was initially lower compared to in-distribution tests, the agent was able to achieve limited success and, most importantly, demonstrated improvement over time (see Figure \ref{fig:out-results}). This indicates that the model can transfer its learning to novel scenarios, suggesting that the meta-learning ability acquired in a restricted domain can generalize to unseen environments. Just as in the in-distribution case, \( \alpha = 0.1 \) significantly  outperforms \( \alpha = 0.01 \). For the rest of our experiments we choose \( \alpha = 0.1 \).

\subsection{In-Context Behavior Stitching}

Humans can compose solutions from individual experiences and acquire expertise in a piecemeal manner~\cite{langley2022computational}. This permits significantly more efficient usage of experiential information. One of the key advantages of the ICRL-trained transformer is its ability to likewise combine experiences in novel ways to solve complex tasks, a phenomenon we refer to as \textbf{in-context behavior stitching}.

\textbf{Setup:} We created two trajectories of the agent walking along paths that cross. One path led to a hole; the other led to a goal. The agent needed to walk part of one path and then switch to the other to get a reward, requiring it to combine multiple skills to solve the game.

\textbf{Results:} The transformer successfully navigated the challenge and received a high reward in all five trials by integrating portions of previous experiences. Figure~\ref{fig:chaining} shows examples of these trials. This demonstrates the model's ability to piece together action sequences that, while not encountered previously in the same episode, result in optimal or near-optimal paths to the goal when combined in a manner consistent with dynamic programming principles. 

This ability to assemble experiences suggests that ICL acquires expertise
in a piece-meal manner, an open question regarding the relationship of ICL to human-like learning~\cite{roberts2024large}.

\subsection{Robustness to Low-Quality Data}

To understand the model's ability to learn from low-quality data with suboptimal actions, we investigated performance when trained on data of varying quality.

\textbf{Setup:} We created several training datasets with different levels of reward quality:
\begin{itemize}
    \item \textbf{High-quality data:} Trajectories from agents that achieved mostly high rewards. We sampled our data, making it five times more likely to select a successful episode that reached the goal than an unsuccessful episode.
    \item \textbf{Mid-quality data:} A combination of high-reward and low-reward trajectories. No sampling weight was given to either successful or unsuccessful episodes.
    \item \textbf{Low-quality data:} Trajectories from random agents with mostly suboptimal actions. We sampled our data, making it five times more likely to select an episode that did not reach the goal than a successful episode.
\end{itemize}

We then trained separate instances of the transformer on each dataset and evaluated their performance on unseen Frozen Lake environments.

\textbf{Results:} One might expect the transformer's performance to degrade proportionally with the ratio of good to bad data as in pure imitation learning~\cite{ghosh2024robust} - a form of supervised fine-tuning. However, as shown in Figure \ref{fig:robust-to-data}, our results indicate that varying the quality of the training data had a minimal impact on the final cumulative reward. Remarkably, the transformer learned effective policies even when trained on mid-quality and even low-quality data, which consisted largely of suboptimal actions and unsuccessful episodes. In fact, training exclusively on high-quality data slightly reduced performance, suggesting that ICRL benefits from exposure to a diverse range of experiences, including both successful and unsuccessful trajectories. This finding demonstrates that the transformer is robust to the quality of the training data and can learn effectively without the need for extensive data curation or filtering.

\definecolor{color1}{HTML}{721817}
\definecolor{color2}{HTML}{2B4162}
\definecolor{color3}{HTML}{FA9F42}
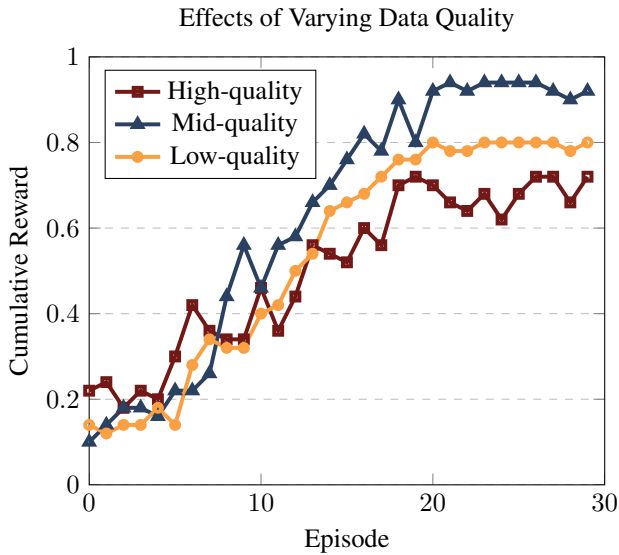
\begin{figure}
    \centering
    \begin{tikzpicture}
    \begin{axis}[
        title={Effects of Varying Data Quality},
        xlabel={Episode},
        ylabel={Cumulative Reward},
        xmin=0, xmax=30,
        ymin=0, ymax=1,
        xtick={0,10,20,30},
        ytick={0,0.2,0.4,0.6,0.8,1.0},
        legend pos=north west,
        ymajorgrids=true,
        grid style=dashed,
    ]
    \addplot[
        color=color1,
        line width=1.5pt,
        mark=square,
        mark size=1.5pt]
        table[col sep=comma,x=episode,y=tau_0.1_high_mean]{data/quality.csv};
    \addplot[
        color=color2,
        line width=1.5pt,
        mark=triangle,
        mark size=2pt]
        table[col sep=comma,x=episode,y=tau_0.1_mid_mean]{data/quality.csv};
    \addplot[
        color=color3,
        line width=1.5pt,
        mark=o,
        mark size=1.5pt]
        table[col sep=comma,x=episode,y=tau_0.1_low_mean]{data/quality.csv};
    \legend{High-quality,Mid-quality,Low-quality}
    \end{axis}
    \end{tikzpicture}
    \caption{Unlike imitation learning, ICRL is largely impervious to data quality. Here we plot the average return achieved over 50 runs when the network is RL-trained on high-quality, mid-quality, and low-quality data. The agent shows significant improvement over time with little dependence on training data quality.}
    \label{fig:robust-to-data}
\end{figure}

\subsection{Adaptation to Non-Stationary Environments}

We tested the transformer's ability to adapt to environments that change over time, reflecting non-stationary conditions that can occur in real-world scenarios.

\textbf{Setup:} We presented the transformer with sequences of environments where the map configuration changed after 30 episodes. The agent was allowed to experiment in the new environment, but was not informed of the change. The changes included alterations in the position of holes, the size of the map, and the start and goal locations. This was repeated 50 times, and the average was plotted in Figure~\ref{fig:nonstationary-results}.

\textbf{Results:} The transformer adapted to the changing environments. It prioritized recent interactions in its decision-making process, effectively disregarding outdated information from previous environments. In Figure \ref{fig:nonstationary-results}, the agent's performance drops when the environment changes and then recovers, rising to almost the same level as previously. The agent learned to adjust its policy based on new information! This adaptability is a key advantage of the ICRL approach, enabling operation in non-stationary environments.

\definecolor{color1}{HTML}{0B6E4F}
\definecolor{color2}{HTML}{989898}
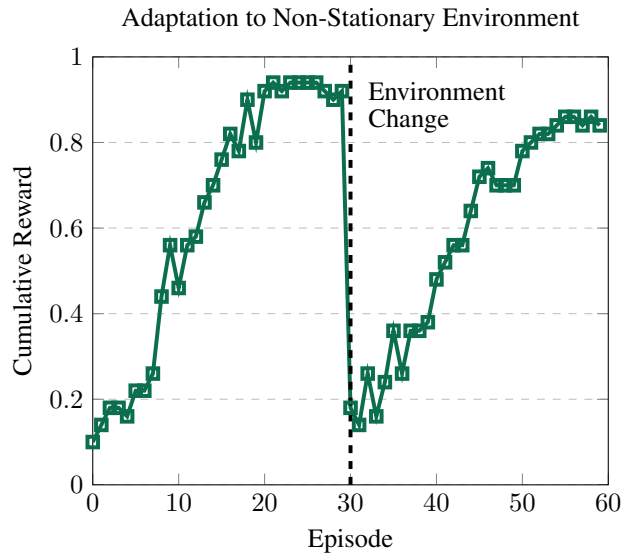
\begin{figure}
    \centering
    \begin{tikzpicture}
    \begin{axis}[
        title={Adaptation to Non-Stationary Environment},
        xlabel={Episode},
        ylabel={Cumulative Reward},
        xmin=0, xmax=60,
        ymin=0, ymax=1,
        xtick={0,10,20,30,40,50,60},
        ytick={0,0.2,0.4,0.6,0.8,1.0},
        legend pos=north west,
        ymajorgrids=true,
        grid style=dashed,
    ]
    \addplot[
        color=color1,
        line width=1.5pt,
        mark=square,
        mark size=2pt]
        table[col sep=comma,x=episode,y=mean_tau_0.1]{data/non_stationary.csv};
    \addplot[
        color=black,
        mark=none,
        line width=1.5pt,
        dashed,]
        coordinates {(30,0) (30,1)};
    \node[anchor=west] at (axis cs: 31,.92) {Environment};
    \node[anchor=west] at (axis cs: 31,.85) {Change};
    \legend{}
    \end{axis}
    \end{tikzpicture}
    \caption{Even when the environment changes, an ICRL-trained transformer can detect and adapt without any explicit signal of the change. The plot shows the average over 50 trials. At episode 30 the environment is changed without any warning to the agent. The agent detects this change and increases its reward through ICRL.}
    \label{fig:nonstationary-results}
\end{figure}

\subsection{The Challenge of Exploration}
\label{sec:exploration}

Despite the strong performance of the transformer model, we observed challenges related to \emph{exploration}. Specifically, during evaluation, if the transformer is not encouraged to take random actions at the beginning of each new episode, it tends to settle into suboptimal trajectories. Even when we enforce exploration, many failures occur because the model has never seen an example of reaching the goal before.

We believe that part of the problem is related to the distributional shift between offline training and online evaluation~\cite{levine2020offlinereinforcementlearningtutorial}. During offline training, the model is provided with a random mixture of successful and unsuccessful episodes. However, at the start of online evaluation, there is a very high proportion of unsuccessful episodes.

The following solutions may address these challenges:

\begin{enumerate}
    \item \textbf{Online Training}: Train the model in an online manner, allowing it to experience low-reward trajectories initially and adapt over time.
    \item \textbf{Model-Based Reinforcement Learning (MBRL)}: Train the model to predict tokens in the environment and roll out experiences based on the chosen actions, effectively simulating online learning.
    \item \textbf{Cross-Episode Reward Function}: Train the network with a reward function where the reward an action receives is based on the expected value of future rewards in \emph{future episodes}. This approach could potentially reward the model for exploration, even if an action does not contribute to attaining the goal in the current episode.
\end{enumerate}

\section{Conclusion}

In this study, we have demonstrated that fine-tuning a pre-trained transformer with reinforcement learning enables it to function as a general-purpose problem solver. It becomes capable of adapting and improving in situations it has never encountered before. While in-context reinforcement learning (ICRL) may not always find the correct answer, the key is that it can \textit{enhance its performance} by adapting in unforeseen scenarios. This progress indicates that agents capable of human-like adaptability and continuous improvement are within reach. 

\appendix

\section*{Ethical Statement}
Extensive analyses have shown that risks to humans escalate as systems become more autonomous; essentially, when users surrender greater control, the potential dangers from the system increase~\cite{AIAgents2025}. These risks are further amplified by adaptable agents that, in theory, can learn to solve novel, unforeseen problems autonomously. Consequently, we propose that agents should not be deployed in unconstrained environments at this time. Instead, we recommend that agents be confined within controlled "sandbox" environments where they are unable to affect the outside world, allowing for rigorous testing and validation of their behaviors.

Some argue that, because of these risks, we should cease the development of autonomous agents (AAs) altogether. However, we contend that the advancement of autonomous agents is both inevitable and essential due to the substantial benefits associated with their capabilities. It is in humanity's best interest to pursue the responsible development of autonomous technologies before malevolent actors possibly exploit them. This approach is further justified by the strong likelihood that the most effective defense against harmful AAs will be the deployment of benevolent AAs designed to counteract malicious activities.

\bibliographystyle{named}
\bibliography{ijcai25}

\end{document}